\documentclass{article}
\usepackage{graphicx}
\begin{document}
\title{An Enactivist account of Mind Reading in Natural Language Understanding}

\author{Peter Wallis\\
Centre for Policy Modelling,\\
Manchester Metropolitan University\\
Business School, All Saints Campus, Oxford Road\\
Manchester, M15 6BH, UK; pwallis@acm.org}

\maketitle

\begin{abstract}
In this paper we apply our understanding of the radical enactivist agenda to the classic AI-hard problem of Natural Language Understanding. When Turing devised his famous test the assumption was that a computer could use language and the challenge would be to mimic  human intelligence.  It turned out playing chess and formal logic were easy compared to understanding what people say.  The techniques of good old-fashioned AI (GOFAI) assume symbolic representation is the core of reasoning and by that paradigm human communication consists of transferring representations from one mind to another.  However, one finds that representations appear in another's mind, without appearing in the intermediary language.  People communicate by mind reading it seems. Systems with speech interfaces such as Alexa and Siri are of course common, but they are limited. Rather than adding mind reading skills, we introduced a ``cheat'' that enabled our systems to fake it. The cheat is simple and only slightly interesting to computer scientists and not at all interesting to philosophers. However, reading about the enactivist idea that we ``directly perceive'' the intentions of others, our cheat took on a new light and in this paper look again at how natural language understanding might actually work between humans.
\end{abstract}

\section{Introduction}
The idea that one should be able to hold a conversation with a computer has been around since the very first computers but truly natural conversation with a machine is still an AI-hard problem.  Enthusiasm for the idea waxes and wanes with a cycle of investor confidence, followed by a burgeoning community of fresh young researchers, followed by discovery, disillusionment, and increasingly limited funding opportunities. From the Alvey Programme~\cite{alvey} and the Microsoft desktop assistant~\cite{clippit}, through the DARPA Communicator programme~\cite{CommEval} to the current Machine Learning approaches, we keep trying but your average 2 year old still holds a more engaging conversation.

The problem turns out to be hard. What we ought to be doing is keeping track of what doesn't work, but this is problematic in a boom and bust environment. It would appear that the current batch of active researchers were never taught the lessons learned last time round and those of us who where about back then all moved on when the funding dried up. Naturally the current batch of NLU champions have a reposte and will claim that a ML approach does not require theory as it treats language understanding/response generation as a black box\footnote{As a reviewer kindly pointed out, there is considerable work being done on addressing this in the interests of explainable AI (XAI). XAI is however addressing concerns of the user; The claim here is that researchers ought to be interested in how these systems work. As stated later in this paper, we embrace the idea that explanations may, indeed should, have nothing to do with how the system actually makes decisions.}.  History is however at play.  Last time round we got quite good at doing a specific thing --- namely task based dialogues, characterised by the DARPA Communicator tasks of booking a taxi, hotels, flights and restaurants.  Twenty something years ago these all looked like near future technology. And they were.  The problem now is that those tasks have become definitive and progress is measured as being able to do these tasks slightly better.  The real challenge is perhaps characterised however by the edge cases in the Communicator data~\cite{Wallis08}, the Alexa Prize~\cite{alexaprize}  and, dare I say it, the Loebner Prize~\cite{loebner}.

What we learned last time was that truly natural language understanding --- conversation as opposed to a spoken language interface~\cite{ibtbagm} --- requires ``mind reading''.  Rather than ``meaning'' being \emph{in} words, people recognise the intentions of their conversational partners and are very good at deciphering what the speaker meant to say rather than what he or she actually said. This is not the ``intents'' used in the Alexa Skills Kit~\cite{ASK} which are what Dennett~\cite{Denn87} wanted to call ``intensions'' (to contrast with the ``extension'' of a sign) but the normal everyday ``folk'' usage that distinguishes between intending to do something and doing it accidentally.  The linguists have known that recognition of intention is key to language understanding for decades but of course the way they say it is not in language that we mathematicians and engineers understand.  What is more, when computer savvy researchers do discover the problem, a suitable solution can appear quite daunting~\cite{trains}.

The premise here is that the apparent difficulty stems from our GOFAI view of what Artificial Intelligence actually is. Some 35 years ago AI researchers discovered that computers could do something other than manipulate symbolic representations of the world, computers can also \emph{do} things in the world. And critically, they can do things without representing anything.  We are all familiar with the notion of a database and this is our collective view of what \emph{real} computers do.  We are also familiar with central heating thermostats, and they are \emph{not} real computers.  As philosophers have pointed out from Wittgenstein talking about bricks, through Austin doing things with words, language in use (as opposed to on the page) \emph{does} things and is thus more like the thermostat than many would like to think.

The point of this paper is to outline a radical enactivist~\cite{enactivism17}
approach to programming a machine to \emph{do} conversation.  The algorithm and
data structures are not interesting and there is no maths. If your interest is
in these then please stop reading now, but for those wanting to make machines
that use language, we hope you will find this paper of interest. The next section gives an overview of the field of Natural Language Processing (NLP) as understood by the expected audience.  The paper then moves on to the problem, and our cheap trick.  We then reinterpret the cheap trick from an enactivist view point, and finally go on to show how the new understanding might be applied to casual conversation.

\section{A popular view of language \& meaning}

The GOFAI approach to language understanding -- the state-of-the-art pre 1990 -- was based on a reference model of meaning in which symbols referred to things (and sets of things) in the world.  The methodology followed what the Soviets called Meaning Text Theory in which the aim was to translate messy natural language into a representation of its meaning~\cite{mel81}. In order to simplify the process, it was split into specialisations with one group of experts working on \textbf{speech recognition} and converting sound into text, another group working on \textbf{morphology}, \textbf{syntax}, and so on through \textbf{semantics}, \textbf{pragmatics} and \textbf{world knowledge} to produce a context free representation of meaning~\cite{all87}.  This would hopefully look something like predicate calculus, but whatever form the representation took, the primary interest in computer science was in the process of transformation.

The focus on translating into meaning left much unsaid about how to make machines hold a conversation.  Dialog systems have tended to focus on question answering and command-based systems because the techniques in NLP are primarily aimed at transformation. To answer a question, a classic approach is to have the computer take what is said, translate that into a data-base query (i.e. its meaning) run the query against the database, and translate the result back into English  -- possibly by passing it backwards through the above meaning-text pipeline. Verbal commands can be treated as translating what is said into meaning where the meaning is a sequence of actions.  There is far more to conversation than question answering but computers answering questions certainly looks like understanding.

We now have a fairly good idea of how to capture the meaning of an English sentence and rewrite it in a canonical form. Rather than wall covering descriptions of English syntax, techniques for \textbf{information extraction (IE)} -- driven largely by the DARPA Message Understanding Conferences~\cite{muc4,muc5,muc6}) -- use database filling gazetteers.  The techniques are data driven, shallow, and result in a representation of meaning that is context dependent.  Having a machine \emph{understand} natural language utterances however requires information from the context.  Whereas speech recognition, morphology and syntax remove information from a utterance and normalise it, pragmatics and world knowledge add information.  Schank and Abelson introduced the idea of ``scripts'' (discussed below) but that approach is quite specific.  The CYC Project may have turned out to be a funding black hole~\cite{wool2020} but the aim initially was to test the idea that such information could be represented in a suitably generic framework~\cite{len94}. Although Lennat may have sold the idea to others who should have been better informed, the initial funding was well spent: the answer is no, we do not know how to create context free representations of meaning.

These days we have a good idea of how to take what is said and extract the information {\em in} text and then use it in a system that provides the contextual information about pragmatics and world knowledge.  For machine translation from Japanese to English for example, if all goes well the human at each end of the translation process have a largely shared context and can fill in the gaps.  Filling in a form to book a flight, the word ``London'' is ambiguous by the old model and might mean the City of London, the bit of the metropolis inside the M25, or the British Government (as in ``London's decision to ...''), or London's Heathrow Airport.  Used to fill in slots in a form for booking flights, the context independent meaning of the term is fixed as planes need airports.  
The relationship with doing is where the action is and for task based dialogs that relationship is reasonably fixed.  For casual conversation -- and indeed the
edge cases in task-based dialogs -- for us social beings who use language to communicate, the doing is embedded in a soup of normative social relationships.

\section{Mind reading}

Language has fascinated us in the West ever since someone claimed that God made the world with the word. There are many different aspects to linguistics, studied in many different ways, and the linguistic theory embraced here is the ethnomethodological variant~\cite{tHave99} of Conversation Analysis~\cite{Sacks, HutWoo98, PomFehr97} (EMCA~\cite{ emca}).  The approach is quite distinct from the classic computer science focus on syntax and parsers, and distinct again from ``sequence analysis'' (see Levinson~\cite[pp 289]{pragmatics} for the comparison)  made popular in NLP by the work on annotated corpora~\cite{swbd-damsl,carCL97}.
The EMCA methodology is to embrace introspection, but in a highly constrained way. The focus is on ``the work done'' by an utterance in context, and the work done is to affect the behaviour of other ``members of the community of practice'' (MoCP).  When the researcher is also a member of the community of practice, he or she (the scientist) can ask him or her self (the MoCP) what work was done by an utterance. The researcher uses inspiration to come up with theory, and uses introspection to test the theory.  As a member of the community of practice, the researcher's plebeian knowledge is, by definition, not wrong - not systematically anyway - and the key is to keep the raw data from introspection separate from theory which must be tested.  The methodology is used later in the paper when we look at chat-room transcripts but here the "findings of CA" are used to refine what is meant by mind reading.  The big picture is that a conversation consists of utterances produced by a speaker and his or her conversational partner (CP) that are broadly sequential and are enacted within a social setting.  At any point in a conversation, the speaker's utterance will go~\cite{Seedhouse04}:
\begin{enumerate}
\item	\textbf{seen but unnoticed} if it is the second part of an adjacency pair.	That is if it is the answer to a question, a greeting response to a greeting and so on. If not, it might go
\item	\textbf{noticed and accounted for} if the CP can figure out why the speakers said what was heard. For example a person walking into a corner shop might say ``Excuse me. Do you sell stamps?'' and the person behind the counter might respond (in the UK) ``First or second class?''.  This is not an answer to the question, but the CP can figure out why the person behind the counter said what he did. If an utterance cannot be accounted-for, the speaker
\item	\textbf{risks sanction}. Basically the CP gets annoyed with the speaker and, depending on a myriad of social factors, will work toward not engaging with the speaker.
\end{enumerate}
The techniques of classic Natural Language Understanding are good at the first of these, but natural conversation is full of the second and, perhaps surprising to many, regularly engages with the third. Abuse is not a product of things going wrong~\cite{wall05-abuse}, but rather a product of the fact our computers cannot recognise the warning signs that their human conversational partners are radiating. Polite conversation between humans surfs a reef of sanction in a way that says an awful lot about power and distance relationships between the participants, and it is something we do without thinking.  What is required with regard to politeness and how to do it is, however, not the issue here.

The issue here is based on the observation that accounting-for is wide spread and indeed essential for participation in naturally occurring conversation. Rather than seeing language as a weak form of formal logic which has its own internal truths, \emph{engineering} a natural language interface embraces the idea of the dialog manager being an interaction manager for a device that interacts with people. How do people go about understanding ``first or second class?'' as a response?  In other fields the study of how one person understands and interrelates with another person has been dominated by two main models of Theory of Mind (ToM), theory theory, and simulation theory ~\cite{gallagher2020}[p69].  Using theory theory, the potential customer in our above example would have a theory about the way other minds work and a theorem resolver that quickly resolves the link from ``first or second class?'' in terms of the shop assistant thinks I want to buy stamps; the shop assistant has stamps; the shop assistant wants to sell me stamps; in order to sell stamps he needs to know how many and what type I want. Hence he has asked me ``first or second class (stamps)''.  Using simulation theory, as the potential customer I would use my own behaviour to account-for the shop assistant's utterance.  By imagining that I was the shop assistant, I can quickly, off line, work through the scenario in which I do not have any stamps, and decide that ``first or second class?'' would not make any sense. Trying the alternative in which the answer to my question was ``yes'', I find that I would think it sensible to ask ``first or second class?''.  Conclusion; the answer to the question I actually asked is ``yes''.  Both these approaches to ToM result in some wonderfully complex algorithms and James Allen~\cite{trains} and others have taken a ToM approach to build fully fledged conversational interfaces.

Accounting-for by reasoning from first principles is expensive and many have proposed there is some ``higher level structure'' to conversations. As an example of the phenomena, Mann uses the following example in his introduction of Dialog Games~\cite{Mann88}:

\begin{tabular}[h]{rrp{6cm}}
  1&Child:& I'm hungry \\
  2&Mother:& Did you do a good job of your geography homework? \\
  3&Child:& Yeah. \\
  4&Child:& What's to eat? \\
  5&Mother:& Let me read it. \\
  6&Mother:& What is the capital of Brazil? \\
  7&Child:& Rio de Janeiro. \\
  8&Mother:& Think about it. \\
  9&Child:& It's Brasilia. \\
  10&Child:& Can I eat now? \\
  11&Mother:& I'll let you have something later. \\
  12&Mother:& What is the capital of Venezuela? \\
  13&Child:& Caracas. \\
  14&Mother:& Fine. \\
  15&Child:& So what can I eat? \\
  16&Mother:& You want some cereal? \\
  17&Child:& Sure. \\
\end{tabular}

\noindent
Notice how there is no semantic connection -- none at all -- between the mother's utterances and what the child says until line 11.  How should we think about what is happening here? More formal logic won't help and although one might argue that it is just pragmatics, the participants are not just talking into the air. Something holds the conversation together.

Of course as English language users we all know what's happening: the child wants to eat; mum wants homework done, and for 20 years the author has advocated that we embrace that folk understanding and provide dialog management systems based on a model of that wanting. The claim is that language in use is cooperative and intentional~\cite{tomasello08}, and in the late 90s several groups looked at using BDI architectures as characterised by Rao and Georgeff's ``Plans and Practical reasoning''~\cite{RaGe95} to manage dialog (see \cite{MaxGuide}, \cite{ArdBoe98}, \cite{wmod01}).  

\section{Representing time-extended action and commitment}

The point of Rao and Georgeff's Beliefs, Desires and Intention (BDI) architecture was not to \emph{do} planning, but to \emph{use} plans in a way that balanced reactive decision making against deliberative behaviour.  The way it does it is to commit to \textbf{time-extended action} (a plan) that is interruptable, and do it based on the system's explicit \textbf{goals}.  Computers are just not fast enough to reason from first principles to atomic actions in what Brooks~\cite{brooks91} characterised as a sense-model-plan-act cycle and instead a BDI architecture chooses an plan and follows it through.  Mean while, it monitors changes in the environment and at any point may stop execution of the plan and choose another.  The choosing turns out to be hard~\cite{wool00}.  The original motivation for using BDI for dialog was the realisation that conversational agents, just like football playing robots, need to deal with a changing world.  The claim now is that the ``higher level structure'' in dialog corresponds to participant goals and, as BDI architectures were designed to reason about action at the level of goals, a BDI architecture is ideal for managing dialog at this \textbf{discourse level}.

BDI architectures and the like are not on the research agenda as they once
were, but that is because the problem has been identified and largely solved.
That said, it is disturbing to see contemporary systems that either use naive
planning architectures that over-commit to a course of action and, at the other
extreme, to see purely reactive systems being used as partial solutions
- partial because a full system would require commitment.  The things committed
to can be ``plans'' as conceived as a sequence of atomic actions, but have also
been called ``activities''~\cite{me04}, are strongly related to ``layers'' in
a subsumption architecture~\cite{brooks91}, and these days are generally called
``behaviours''~\cite{Ark98}. The point is not the mechanism, but the
commitment.

Conversation as a sequence of actions is an extremely popular way to represent time-extended action in dialog management systems.  Recent examples include the ``dialog designer's work bench'' for Samsung's Bixby~\cite{bixbyDev} and Skantze's equivalent for the furhat~\cite{furhat}.  State machines have nice theoretical properties but, as an unaugmented dialog control mechanism, over commit. Once a conversation has settled on a path through the network, the only way for a user to change his or her mind is either to have had the dialog designer put in an explicit ``jumps'' or to ``start over''. Inevitably the hard part of the problem is left to the research assistants.  State transition diagrams are however an intuitive way to represent action and were how plans were represented in our original work on BDI dialog systems~\cite{wmod01}.

The ubiquitous chat-bot mechanism is another mechanism for providing time extended action, and one that does not over commit.  In pure form as represented by the original ELIZA, it does not over commit because it does not commit at all.  If one says ``cheese burger'' in the middle of a conversation about booking a flight, a human will work hard to \emph{account-for} the utterance whereas a chatbot cannot tell that such a statement is out of context. Contemporary augmented chatbot mechanisms of course collect information and are extremely common for task based dialog on web-sites. Indeed the Alexa Skills Kit for the Amazon Echo is essentially using the same technology. Rather than something that looks like a plan from classic AI, the interactive nature of a conversational machine means something more like task-based chatbots (tagged with the goal each might achieve) can be used in place of plans in a dialog system using a BDI architecture.

The VoiceXML standard introduced forms as a representation of time extended action in dialog. The aproach was to use XML to represent a form with ``slots'' to be
filled. Each slot can have a set of IE style rules that say how the slot can be
filled from what the user says.  A slot for ``destination'' might for example
have rules saying how to map user utterances into airports so that, if, the
user says ``I want to go to London from ..'' the destination slot will be
filled with ``LHE'' representing London Heathrow airport. Each slot can also
have a set of ``prompts'' for the system to use when it comes time to speak
(and the slot has not been filled). For the destination slot, the first prompt
might be ``Where do you want to fly to?''.  Note how the system allows ``mixed
initiative'' in the sense that the user does not need to wait for the prompt
before providing information. The user might say ``I want to fly to London next
Tuesday'' and the slot fill rules can pick up both the destination and the
departure date.
Once again, deployed Voice XML systems tend to over commit in that, once the
system has sellected a form to fill, the only way for the conversation to move
on is for the form to be completed or for the dialog designer to have put in
recognition rules that link to other forms -- links much like hyperlinks in an
html page.  The result is that often the only way for the user to change his or
her mind is to ``start over''.

\begin{figure}
\includegraphics[width=\linewidth]{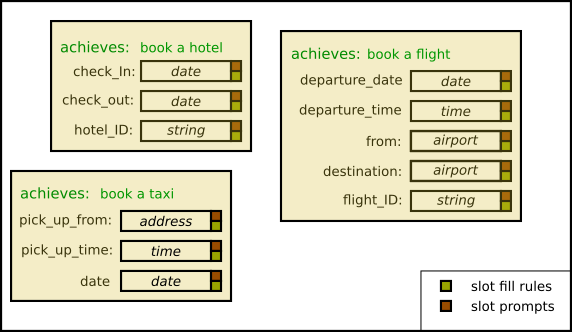}
\caption{A "plan" library based on VoiceXML style forms where slots have
associated slot fill rules (Information Extraction) and prompts.} \label{L1}
\end{figure}
VXML forms is our preferred way to represent time-extended action in a dialog
system. The real issue however is for a system to decide which time-extended
action to commit to and when.  Figure~\ref{L1} represents part of
a plan-library for the Communicator callenge of booking flights, hotels and
taxis.  The challenge for a dialog management system is to choose which form to
use.  The challenge for the dialog designer is to choose which forms to add to
the library. The apparently controversial claim is that what constitutes a form
-- what makes a set of slots belong together -- should be based on the
percieved goals of the user. For task-based dialogs this approach does seem to
be fairly hard to contest although it is interesting to look at what people
actually try to do when using a system~\cite{Wallis08}.  A goal based analysis
is certainly harder for casual conversation but techniques from linguistics
such as CAEM can certainly shine a light on the darker corners of human
behaviour. The point of this paper however is to do as the author's betters
wanted and to bannish goals from the model. In order to do that, the next
section describes how one might use a library of goal-based time-extended
actions if one could read minds.

\section{The wrong approach: {BDI} with intention recognition}

When it comes to having a conversation with a machine, we humans expect the machine to commit to time-extended action and the claim is that the required ``high level structure'' is goal based.  The problem with the classic chat-bot is a reactive system that can't commit, and a frustratingly large number of well resourced dialog systems over commit. 

The classic way to use a BDI architecture is top down with the system having
a goal, then finding a plan from the plan library to execute.  As the environment changes, the system checks the current plan for appropriateness and possibly initiates a new plan. If we had a system with a mind reading module, the classic way to use a BDI architecture for dialog would be to give it a plan-library like that in Figure~\ref{L1} and use the following algorithm:
\begin{verbatim}
while engaged in a conversation with the user
  wait for the user to stop talking or a time-out
  read the user's mind and get userIntention
  check the goal achieved by the current plan aligns userIntention
  if it doesn't
    choose a new plan
  say the next thing from the current plan
\end{verbatim}

Instead of creating a mechanism for reading minds based on the current theories
about Theory of Mind, we introduced a simple mechanism that approximates this for task based dialogs.

\subsection{A Cheap Trick}
The general idea for our cheap trick for intention recognition is that people do not do things randomly and so, if a conversational partner does something that does not progress the current plan, then she is probably doing something else. If what she does is part of another plan, then that plan is likely to be what she \emph{wants} to do.

In order for the mechanism to work, the agent needs to know what other plans its conversational partner might have.  An observation was that for task based dialogs there is no point a system recognising the intention of the user if it does not have a plan for dealing with it.  It may be nice to be able to say "I am afraid I cannot book a hotel for you" but being able to say "I am afraid I do not know how to do that" is equally acceptable. A machine only needs to know what it can do, not what it can't. This of course is represented in our systems by the plan library.

The key observation for our poor man's approach to intention recognition is that when things go well, progress is made and slots are filled.  Our cheap trick was to check that each user utterance filled at least one slot in the current form (or did some other action) and if not, then there was considered to be ``trouble in text''.  The system would then set out to \emph{account-for} the user's chosen utterance.  Say "cheese burger" when booking a flight, and our system would notice, attempt to account-for and, if necessary, work toward sanction.  The way it tried to account-for was to look for another behaviour (another VXML style form tagged with a goal) in the plan-library that could do something with what was said.  If something was found, the system assumed the user had changed his or her mind and was now trying to (get the system to) do the goal associated with the new form. If, part way through booking a flight, a caller talks about the Waldorf Hotel, and the behaviour for booking hotels would become active and the system can say ``you want to book a room at the Waldorf - okay."  The system might not be reading minds but it could certainly take a good guess at the user's goal -- and it could do it without representing other minds.

The new algorithm is:
\begin{verbatim}
while engaged in a conversation with the user
  wait for the user to stop talking or a time-out
  check the latest utterance contributes to the current plan
  if it doesn't
    look for a plan in the plan-library that it does contribute to
    if found
      say "Oh, you want to <achieves goal statement>"
      make it the current plan
    else
      say "Um..."
      make the plan to find out what the user wants the current plan
  else
    say the next thing from the current plan
\end{verbatim}
Of course the system might not have a behaviour that recognises the user's utterance but this happens all the time in human-human dialogs.  The brief answer to addressing this issue is to run Wizard of Oz (WoZ) style trials and record how expert humans negotiate the social relations, then to use EMCA to elucidate the strategies used and add them as behaviours to the system's plan-library. As happens in the Mother and Child dialog, those strategies may have very little to do with semantics and quite a bit to do with goals and social status.

The approach can be criticised and improved in many directions but the point is
to introduce a framework that emphasises the way humans negotiate, not shared
knowledge, but shared goals.  There are of course some obvious issues.
First, the issue of coverage which plagues all dialog systems based on hand
crafting rules.  It turns out that humans have a notion of fairness and what is reasonable~\cite{tomasello08} and thus it seems quite feasible that, for task based dialogs at least, a dialog designer can give a system ``coverage''.  Having given a system all the behaviours to handle that which might reasonably be expected of the system, the system is then socially justified in moving toward sanction if the user strays ``off topic''. In Wizard-of-Oz trials in 2000 we had our wizard giving directions around campus. In what Garfinkel~\cite{garf67} called a breaching experiment, one of the researchers rang and asked to book a table for two. The wizard's response was to say ``You have called the University of Melbourne. How can I help?'' This is not an apology for not being able to help. It is a call on the authority of the institution and a threat to hang up.

A second point worth making is that people assume an intentional stance~\cite{Denn87} when talking with others and plans and goals are regularly treated as first order objects. We do not just make and use plans, we talk about them.   In another set of WoZ trials~\cite{wmod01} our wizard was continually saying things like ``I'll just book that for you; is that okay?'' She said what she was going to do and is ``putting it on the table'' to be checked, modified, and authorised.  The process of managing the conversation is transparent with not only the current behaviour but the entire plan library up for discussion.  If the system's guess about the user's intention is wrong, it does not matter as the user will say. Or, indeed, the system can ask. If a user starts to book a hotel before the flight booking is completed, the system could say something like ``Can I just $<goal>$book a flight $</goal>$ for you first.  Do you have your passport number there?''. This may not be what the user wants, but giving the system access to (descriptions of) goals enables the system and the user to talk about it.

The missing ingredient from conversational user interfaces is mixed initiative at the discourse level and, for us mere humans, ``goals'' seems to be the right way to think about what ``discourse level'' actually means. Recognising the goals of one's conversational partner would seem to be a core part of the process and our cheap trick mimics that skill to a significant extent for task-based dialogs.  However, in the last year or so, we have come to believe that our poor mans version of intention recognition might indeed be how things actually work. What is more, the theory behind this realisation suggests a way to make machines that do casual conversation.

\section{A Radical Enactivist Approach}

The philosophers are considering the idea that we ``enactively perceive (rather than infer or simulate) the intentions and emotions of others.''~\cite{gallagher2020}[p121]. The relevant literature is voluminous and hard-going for us engineers, but the above gives a personal background from which the following interpretation of ``enactive perception'' is derived.

Consider a sheepdog watching a flock of sheep.  One sheep leaves the flock and heads toward an open gate. The classic interpretation would be that the dog sees that the sheep is trying to escape and thus, not wanting the sheep to escape, initiates action to head it off.  Before the dog can get between the sheep and the gate, the sheep recognises the intention of the dog and turns back.  The dog reads the mind of the sheep; the sheep the dog, and issues resolve without bloodshed.
The equivalent for the mother and child example is that the mother recognises the intention of the child; the child, the mother; and a shared (temporary) goal is negotiated. In order to do this with the dialog manager described above, the system needs to recognise the intention of the child/mother and hence find the right behaviour to use based on the goal. Our cheap and cheerful trick enables us to do that to some extent.

The alternative interpretation is based on a behaviour-based robot architecture~\cite{Ark98} in which sensing and acting are connected in layers.  The sheep leaves the flock at velocity (a speed and direction) producing a ``trajectory'' that the dog can see.  The trajectory itself is recognisable.  Rather than recognising a sheep and then doing it again and seeing that the sheep has moved, the dog sees movement and texture that ``brings on'' actions to intercept.  The dog may then choose, somehow, to intercept or not.  The intercept behaviour connects sensing to acting and ``triggers'' of its own accord, without volition. The dog however chooses to enable or inhibit the behaviour. The dog's planning (such as it is) commits the dog to a time extended action by choosing between (triggered) behaviours.  The dog chooses to intercept; the dog's intercept behaviour then takes over.

It is that behaviour that is directly perceived by engaged agents such as other dogs, sheep and indeed shepherds.  As the dog starts to move, the sheep remembers the terror it felt from the last time a dog looked it in the eye and remembers the worrying sensation of having a dog up close and personal. The sheep reconsiders its escape trajectory, and makes a choice: to continue with the dash for the gate behaviour, or activate the return to the flock behaviour...

The farmer of course is different. Whereas the dog derives pleasure directly from having all the sheep in one place, the farmer wants them there for shearing.  People can, and do, do means-ends reasoning but the radical enactivist line is that they do it with the symbolic representations reflected in language.  There is not symbolic reasoning ``all the way down'', and there is no need for ``mentalese''.  The shepherd might \emph{explain} her dog's behaviour in terms of sheep and gates but, these undeniably symbolic representations are grounded in human behaviours.  The notion of a sheep is grounded in the use-cases of herding, chasing, shearing, drenching, and butchering.  Goats may be genetically distinguishable but our ancestors didn't know that. The reason there is a separate word for each is that a goat does not fit all the use-cases for a sheep. The meaning of the symbols we use to represent the world is not grounded in sense data, nor in our relationship between the body and the things around it~\cite{ruthrof97}, but in embodied doing.

Our cheap and cheerful BDI dialog management system recognises the behaviour of the human by recognising that the system itself has a behaviour that \textbf{meshes} with that of the user. Like a pair of gears the action of each tooth ``brings on'' the next. Rather than recognising goal failure, we simply identify when the meshing fails. Goal like linguistic descriptions attached to each behaviour are very handy for talking about these behaviours, but need not be part of the mechanism.

Returning to the mother and child dialog, yes the child does need some
motivating state representing hunger, but that goal does not need to be symbolic
as it is internal to the agent. What is more, viewing the interaction as a
meshing of behaviours we do not need to represent the goals of the mother and hence no need to read minds.  Working through the process, consider the case where the child is hungry -- has the (internal) goal of eating something -- and remembers an interaction that went like this:

\begin{tabular}[h]{rrp{6cm}}
  1&Child:& I'm hungry \\
  2&Mother:& You want an apple? \\
  3&Child:& Sure. \\
  4&Child:&$<$eats apple$>$\\
\end{tabular}

\noindent
This was a multi agent trajectory in which the mother's behaviour ``meshed'' with the behaviour of the child and resulted in no longer being hungry.  In the original mother and child example the child initiates the same behaviour with ``I'm hungry'' but mother does not produce behaviour that meshes.  The child has failed to achieve his goal. He could walk off (and has no doubt tried that before) but instead finds a behaviour that meshes with mother's behaviour.  There is no semantic connection because that is not the level at which this negotiation occurs. Like two sheep running for the same hole in the fence, the behaviours of the agents simply ``bump up'' against each other. Rather than the child reading the mind of the mother and identifying her goal, and mother recognising the goal of the child, each agent has his or her own goal and a behaviour that, all being well, meshes with the other agent's behaviour to form a multi agent trajectory. When things bump, the agents try other behaviours.

Of course attaching a \emph{description} of a goal is useful, not for reasoning about system behaviour, but for \emph{explaining} system behaviour to us poor mortals that \emph{talk} about plans and goals, even if we don't use them to produce everyday behaviour.

\section{Casual conversation}
The epiphany was that our cheap trick \textbf{does not need the goals}.  That is, in Figure~\ref{F1} we don't need the green bits.  If the user is (behaving as if he wants to be) booking a taxi, then the matching ``plan'' in Figure~\ref{F1} will ``mesh'' with the user's behaviour without the system knowing that he wants to book a taxi. If he is not behaving as if he wants to book a taxi, that plan won't mesh. Rather than recognising the intentions of others, all we need is to recognise when a behaviour of our own meshes with the actions of others. Goal descriptions might be useful for talking \emph{about} behaviours, or indeed for reasoning about behaviour at some abstract level, but they are not part of the mechanism.

Having banished all but a vestige of goals from our mechanism for task based dialog, can the mechanism be applied to casual conversation? To this end we took a look at the NPS corpus of chat room conversations available through the NLTK project~\cite{nltk}. The aim was to see if multi agent trajectories could be used to explain the phenomena observed in the data.

As mentioned above the methodology used is that of EMCA in which the evidence is provided by the researcher's inner ``member of the community of practice of English speakers''. The reader, also, has access to an expert in the usage of English and, as part of the methodology, is invited to consult said expert on the quality of the claims made. The reader is however asked -- as part of the methodology -- to leave scientific theory at the door. If you would not expect the subjects to understand a statement about the data, it's theory.

Looking in the 11-09-adults corpus, there are roughly 100 users but about half are lurking and say nothing. About 20 enter the chat room and say ``hi'' or another greeting that offers no information. My plebeian knowledge leads me to hypothesise that saying ``hi'' will elicit a greeting-response of some kind.  Looking at instances of ``hi'' (and variants) I test the theory and find this does not happen in most cases. The theory is wrong. It seems my consultant English using expert -- my inner pleb -- is not a member of this particular community of practice.  He is not alone however and User33, after saying hi and waiting a bit, types: ``thanks for the no hello'' and leaves the chat room.  A key point here is that both User33 and I are indeed members of a broader community of English users and indeed would be perfectly capable of participating in this chat room community with a bit of effort. \emph{How} we would go about it is indeed highly interesting but again not the point here.  The point here is that, like the child expecting mum to say what's to eat, those 20 users are using a behaviour they expect will elicit a response that ``meshes''.

Others try different openings.  User20, out of the blue, says ``base ball is my favourite sport!'' to which User22 responds with ``sleeping is my favourite sport'' and the trajectory about base ball is abandoned. User11 tries ``where from User12?'' which brings on an equally dismissive ``somewhere in the USA'', and User55 tries ``hi how is everyone tonight'', which elicits no response. As Tolstoy notes in Anna Karenina, they are failing to find a ``topic that would stick''.  User100 opens with ``37 m wv''  which, in chat room speak, means Male, aged 37, from West Virginia. User100 seems to have done this before and his opening statement brings on a response.  User22 is highly active and responds to this with ``User100 what part of wv are you in?''.  It seems User22 has a behaviour for talking about West Virginia.  The behaviour has been activated; she has had the chat room equivalent of a silence in which neither she nor User100 is expected to talk; and she performs the first prompt in the enabled behaviour.  That brings on User100's response of ``beckley'', which does not fit with User22's behaviour. The plebeian me thinks she was \emph{want}ing to participate in the pleasurable (internal goal) conversation of reminiscing about shared experiences or gossiping about shared acquaintances. It did not work and the behaviour is dropped. But she has another that opens with the offering of a piece of information.  She says ``ah ok.. my dad lives in clay wv''.  User100 does not have any behaviours that deal with Clay, West Virginia, and User22 abandons the trajectory.

Some trajectories do take hold however.  In Figure~\ref{F1} User27 and User29 are lovers.  The point is that these multi agent trajectories only take hold if other participants are willing to play.
\begin{figure*}
\footnotesize
\begin{tabular}[h]{p{50mm}p{110mm}}
User29!!! & \\
 &				.3(((((..6 User27 ..3))))) \\
 &				User27! \\
 &				.3(((((..6 User27 ..3))))) \\
..6(((.3 User29 .6)))  & \\
 &				User27!!!! \\
 &				I love you~! \\
 &				And i missed you!!! \\
I love you too & \\
I missed you yesterday & \\
 &				I can't stay on to late tonight I have homework =( \\
shane told me i couldnt call you & \\
i was sad & \\
 &				LOL Why couldn't you call me?? \\
 &				my internet was down till this morning \\
 &				.ACTION waits. \\
shane told me not too..so i didnt & \\
i dont have school tomorrow & \\
hahahaha!!!!!! & \\
 &				lmao he told you not to call me yesterday \\
its a holiday!!!!!! & \\
yup yup & \\
lol & \\
 &				LOL I have clinicals \\
 &				Well don't listen to him call me when you want too \\
i think he was joking & \\
 &				My aunt isn't doing well \\
but i didnt chance it & \\
lol & \\
oh no & \\
that sucks hon & \\
 &				LOL you call me whenever you want too \\
 &				Yes she not waking up from surgery \\
 &				:( \\
oh no :( & \\
 &				.ACTION is scared. \\
i hope she wakes up & \\
and gets better & \\
brb & \\
 &				I know to much about stuff like that now... \\
 &				and it scares me \\
 &				me too \\
ok & \\
im back & \\
 &				Welcome back User50 \\
 &				LMAO \\
 &				User27 \\
 &				:p \\
lol & \\
you tabbed wrong!!!!!! & \\
ahhahahaha!!!! & \\
 &				.ACTION trying to keep me spirits up. \\
 &				LOL I know User27 \\
 &				They want let no body but immediate family \\
 &				back there ... makes me sad \\
$<$interacts with others$>$ & \\
 &				brb got to go here my songs \\
\end{tabular}
\caption{Two lovers chat in the NPS data.\label{F1}}
\end{figure*}
There is of course information in this transcript, but the information has very little to do with how participants produce the transcribed utterances. Instead the proposal is that there are identifiable multi party trajectories in this transcript that other members of the community of practice will not only recognise, but also be perfectly capable of using.

Interestingly, note how the chat room interface allows multiple trajectories to be pursued simultaneously. In the usual case trajectories are negotiated, not by semantics, but by simply ``bumping up against each other''.  In a phone call to book a flight, there modality means flights are booked and \emph{then} hotels or a taxi. Trajectories are engaged sequentially. In the mother and child conversation above, mother presented the idea that getting something to eat was dependent on doing homework first.  That is, for some reason, the two trajectories were mutually exclusive, and homework came first.  In the lovers conversation trajectories continually overlap, with apparent contradictions. After the introduction of the subject of the sick aunt User27 says  ``lol''(laugh out loud) and ``oh no'' in the same turn, but these are correctly ascribed to different, parallel, conversational trajectories. There is no ``bumping'' and so no negotiation is required.

\section{What next?}

From the perspective of good old fashioned AI, trajectories have much in common with Schank and Abelson's notion of scripts. Scripts were higher level data structure that captured the pragmatics and world knowledge of a situation in such a way it could be used in natural language understanding.  As Shank later said~\cite{schank89}, these were meant to be ``frequently recurring social situations involving strongly stereotyped conduct'' but were seen by others as something more generic. The problem Schank had was to get a machine to map from a particular instance of a story about going to a restaurant to the generic template for conduct in a restaurant. The solution at the time was to produce abstract representations using Conceptual Dependency graphs. This approach - to represent things and events in the world with symbols in a machine - is classic GOFAI and so, although in theory CD could provide a solution, it is not practical.

The exciting possibility is that the trajectories humans use are actually quite shallow personal ``narratives'' that can be extracted from WoZ transcripts.  Rather than needing to be ``strongly stereotyped conduct'' so that hand crafting scripts is manageable, or so that ML techniques have sufficient training data, behaviour in social situations can simply be extracted from single cases in the corpus. Rather than abstracting from ``forgot my wallet'' to some representation of the idea that restaurant goers need to pay for their meals and paying involves a wallet and so on and then finding similar cases that have also been abstracted in the same way, the idea is to remember the embarrassment in the context of a set of enabled behaviours.  Those behaviours might include making a run for it, borrowing money, and letting the restaurant owner figure it out and agreeing to everything he says. Having washed dishes in order to pay for a meal need only happen once to have the event indexed by the presence of dirty dishes and suitable embarrassment. Indeed, the agent does not even need to conceptualise the doing of dishes in order to say ``I'll do dishes for you'' and be presented with a washcloth.   Implementing this on a computer is beyond us at the moment and we expect more work needs to be done understanding what the philosophers mean by ``primary and secondary inter-subjectivity,'' and that is work in progress.

In the mean time, instead of mining for behaviours from corpora - corpora we do not have yet - we have been working on a ``dialog designer's work bench'' that, like the Alexa Skills Kit, AIML~\cite{alice}, and VoiceXML, enable someone with the skills to author scripts for human-machine dialog.  Our interest is in casual conversation and for various reasons the specific interest is in story-telling.  SINDL - Spoken Interactive Narrative Description Language - is a JSON based approach to writing interactive books that can be read by a machine.  This is certainly possible with existing systems such as the Alexa Skills Kit but, coming from the tradition of meaning text theory, skills such as ``The Secret Door'' are inevitably question and answer based. By having a notion of back-channel expressions, and mixed initiative at the discourse level, we believe we can have far, far, more fluid interaction.

\section{Conclusion}

This paper outlines an algorithm for ``enactivly perceiv[ing ...] the intentions [...] of others''. The reason us engineers ought to be interested is because, as the linguists point out, there is more to dialog than the computer science notion of ``understanding'' natural language.  Clever algorithms from Machine Learning \emph{might} deliver the goods one day, but the Amazon Echo has been about for a while now and the claim by ML advocates that they just need more data or more computing resources to crack the problem is becoming tired.

The claim here is that the primary building block of language is the behaviour.  Behaviours interface volition with the world, and we talk about (the goals associated with) behaviours. We do not identify things in the world and then think about them with language tacked on top; we engage with the world and then use language to engage with others. If others behave in a way that meshes, we begin to suspect that our words mean something.

Our next step is to build something that works. The techniques of EMCA might be convincing, but the author's inner engineer really want to make a machine that can hold a conversation.

\bibliographystyle{SageV}

\end{document}